\documentclass{sig-alternate-05-2015}

\usepackage{amsmath,amssymb,bm,dsfont,enumitem} 
\usepackage{color}
\usepackage{xfrac}
\usepackage{epstopdf}
\usepackage{algorithm}
\usepackage[noend]{algpseudocode}
\usepackage{relsize}
\usepackage{booktabs,bigstrut,multirow,multicol,makecell}
\usepackage{epstopdf}

\newcommand{\todo}[1]{} 
\newcommand{\nir}[1]{} 
\newcommand{\elad}[1]{} 

\newcommand{\expect}[2]{\mathbb{E}_{#1}\left[{#2}\right]}
\newcommand{\inner}[1]{\langle {#1} \rangle}

\newcommand{\R}{\mathbb{R}}

\renewcommand{\P}{\bar{P}}

\newcommand{\cX}{{\cal X}}
\newcommand{\cXp}{{\cal X}'}
\renewcommand{\c}{\mathsmaller{C}} 
\renewcommand{\t}{\mathsmaller{T}} 
\newcommand{\D}{{\Delta}}
\renewcommand{\d}{\mathsmaller{\Delta}} 

\newcommand{\cYc}{{\cal Y}_{\c}}
\newcommand{\cYt}{{\cal Y}_{\t}}
\newcommand{\proxy}[1]{\tilde{#1}}

\begin{document}

\setcopyright{acmcopyright}





%

\title{Predicting Counterfactuals from \\Large Historical Data and Small Randomized Trials}

\numberofauthors{3}
\author{
\alignauthor
Nir Rosenfeld\\
       \affaddr{Hebrew University of Jerusalem and}\\
       \affaddr{Microsoft Research}\\
       \email{nir.rosenfeld@mail.huji.ac.il}
\alignauthor
Yishay Mansour\\
			 \affaddr{Microsoft Research and}\\
			\affaddr{Tel Aviv University}\\
       \email{mansour@microsoft.com}
\alignauthor Elad Yom-Tov\\
       \affaddr{Microsoft Research}\\
       \email{eladyt@microsoft.com}
}

\maketitle
\begin{abstract}

When a new treatment is considered for use,
whether a pharmaceutical drug or a search engine ranking algorithm,
a typical question that arises is, will its performance exceed that of the current treatment? The conventional way to answer this counterfactual question is to estimate the effect of the new treatment in comparison to that of the conventional treatment by running a controlled, randomized experiment. While this approach theoretically ensures an unbiased estimator,
it suffers from several drawbacks, including the difficulty in finding representative experimental populations as well as the cost of running such trials. Moreover, such trials neglect the huge quantities of available control-condition data which are often completely ignored.
 
In this paper we propose a discriminative framework for estimating the performance of a new treatment given a large dataset of the control condition and data from a small (and possibly unrepresentative) randomized trial comparing new and old treatments. Our objective, which requires minimal assumptions on the treatments, models the relation between the outcomes of the different conditions. This allows us to not only estimate mean effects but also to generate individual predictions for examples outside the randomized sample.
 
We demonstrate the utility of our approach through experiments in three areas: Search engine operation, treatments to diabetes patients, and market value estimation for houses.
Our results demonstrate that our approach can
reduce the number and size of the currently performed 
randomized controlled experiments, thus saving significant time, money and effort on the part of practitioners.

\end{abstract}


\section{Introduction} \label{sec:intro}

Novel medical treatments, new government policies, and innovative website designs are all examples of changes to an existing method of interaction with people that need to be evaluated for their effectiveness before they can be put into use. The gold standard for testing such interventions are randomized controlled trials (RCTs) \cite{rcts}. RCTs are widely used in medicine: Approximately 200,000 RCTs were conducted in the 1990's alone \cite{rcts}. Internet website operators were early adopters of RCTs \cite{kohavi2009controlled}. Most large Internet companies are known to run thousands of RCTs every year \cite{kohavi2012trustworthy}. 

RCTs work by randomly assigning every subject to either a control group or a treatment group.
The average measurement of the result variable for each group is then an unbiased estimator
of its corresponding population mean.
Given these, unbiased estimators of other desirable quantities such as
the mean treatment effect can be easily constructed.

This approach, while appealing, has several drawbacks.
First, for the estimators to be truly unbiased, subjects must be sampled i.i.d.
from the general population of interest.
Not only is this unrealistic and seldom the case, but often times the sample
represents a very specific sub-population, which can lead to extremely biased estimates.
This is especially evident in clinical trials, where subjects 
(who typically volunteer to take part in an experiment)
are often those suffering from severe symptoms,
those which no other treatment helped,
or simply those who are more prone to volunteer.

Second, as controlled trials are expensive and time consuming,
sample sizes tend to be small. This greatly limits the amount of information
available to researchers and practitioners for drawing conclusions,
generating predictions,
and deciding on policies.
The small samples are typically sufficient
for constructing estimators with reasonably low variance,
but are seldom enough for generating high-accuracy predictors.
For instance, in search engine A/B tests,
the decision of whether to use an alternative results ranker
(or even whether to continue running the experiment)
is often based on the average measures of click-through rate (CTR) or similar measures,
and not on predictions regarding specific queries.
Larger samples should potentially allow for the application
of high-end learning algorithms.

Third, the price paid for guaranteeing that the estimators are unbiased
is that only data from the controlled trial can be used.
This completely discards the huge quantities of
data that are often times available for the control condition,
which in most cases is just the current policy.
For instance, consider the case of predicting whether 
administering a new drug would prove better than
the current standard for a given patient.
A predictor trained only on the results of
a small-scale clinical trial should prove to be inferior to one
which also takes into account
all the past medical records corresponding to the
currently applied drug.
Using only the trial results seems wasteful in terms of data,
and generally suboptimal for prediction.

Given the above, the question we pose here is the following:
how can we design a learning algorithm 
for generating predictors in counterfactual settings,
which takes as input both
a small randomized trial dataset,
and a large labeled dataset of the control population?
Answering the above question is the motivation behind this paper.

Note that although the setting we discuss is of a counterfactual nature,
our goal is in essence a predictive one.
In pursuing the goal of generating high-accuracy
predictors, we knowingly forfeit the ability
to explain the underlying causal mechanism.
The latter objective has been the focus of an abundant 
body of works, most based on the framework of
\emph{causal inference} \cite{pearl2009causality}.
We argue here that there is an inherent
tradeoff between interpretability and
predictive performance,
and that when the goal is to optimize accuracy, a direct approach is preferred.
Our work follows the more recent line of work
where a  discriminative
loss-centric approach is applied
in counterfactual settings
\cite{swaminathan2015counterfactual,swaminathan2015self,johansson2016learning}.

The paper is organized in the following manner.
We begin by covering related material in Sec. \ref{sec:related}.
We present notations our and problem statement in Secs. \ref{sec:notations} and \ref{sec:problem},
respectively.
Sec. \ref{sec:method} contains a detailed description of the core of our approach,
followed by Sec. \ref{sec:extensions}
in which several extensions are presented.
Sec. \ref{sec:experiments} contains several experiments
on real data.
We conclude with a discussion in Sec. \ref{sec:discussion}.

\section{Related material} \label{sec:related}
Our setting draws relations to several lines of work.
The fundamental property of the prediction task we consider
is that it is counterfactual in nature.
Causal inference \cite{pearl2009causality} is a standard
framework for estimating the causal relation between variables,
in a way which can then be used to answer counterfactual questions.
In order to achieve this, methods for causal inference
are usually based on simple, interpretable models from which 
actionable conclusions can be drawn
\cite{bottou2013counterfactual}.
Our approach is different in that it focuses on 
prediction by introducing an ad-hoc loss function for parametrized predictors.
Classic causal-inference models on the other hand
do not always allow for arbitrary features,
nor is it always straightforward to learn or to
generate predictions from a given model.

Alternatively, counterfactual questions can be answered if data can be collected under a random policy
\cite{li2010contextual,bottou2013counterfactual,li2014counterfactual}.
In practice, true randomization is difficult to obtain for business reasons (in the Internet setting) or for ethical reasons (in the medical and social domains). Because of this, it has been proposed to treat data collected under different settings as randomized, and use it to answer counterfactual questions \cite{li2015toward}.

More recently, notions from causal inference have been
incorporated in to discriminative learning methods,
with the declared goal of minimizing loss.
In analogy to Empirical Risk Minimization,
the principle of Countefactual Risk Minimization
is proposed in 
\cite{swaminathan2015counterfactual,swaminathan2015self}.
The proposed method offers a discriminative learning objective based on inverse propensity scores,
where the variance is controlled by a regularization term or by self-normalization.
In contrast to our setup, this method requires
that in addition to examples $x$ and labels $y$,
each sample must also includes its loss logged propensity score.
Other works use doubly-robust methods which are based
on propensity scores as well \cite{bang2005doubly}.
Some parametric non-linear methods for estimating treatment effect
are based on Bayesian regression trees \cite{chipman2010bart}
and random forests \cite{wager2015estimation}.

A parallel discriminative approach to 
counterfactual prediction is based on the notion of
domain adaptation \cite{bendavid2010theory}.
Following the work of \cite{scholkopf2012causal},
the authors of \cite{johansson2016learning}
observe that generalizing from the observed factual distribution
to the unobserved counterfactual distribution
is a special case of covariance shift,
and in general of domain adaptation.
Therefore, the non-convex representation learning method in \cite{johansson2016learning}
incorporates a discrepancy-based regularization term which encourages a
label-invariant representation.
In contrast, our method regularizes the relation between the 
control and treatment variables themselves, conditioned
on a the original shared representation.
Moreover, while the method of \cite{johansson2016learning} requires large
amounts of data for both labels, our method is tailored
for a setting where the treatment variable is rare.

%
%

\section{Notations} \label{sec:notations}
Our setup is similar to a standard supervised learning setup
where we are given a sample set of examples $x$ and labels $y$,
but with some additions.
We assume examples are from a general domain $\cX$,
and denote by $\cXp \subseteq \cX$ the sub-domain
of examples that take part in the controlled trials.
Our setup includes two label domains, denoted by
$\cYc$ for the control variable
and by $\cYt$ for the treatment variable.
Throughout the paper we use the terms label, variable, and experimental outcome interchangeably.

Instantiations of examples are denoted by $x \in \cX$,
and of labels are denoted by $y_\c \in \cYc$
and $y_\t \in \cYt$.
We assume there exists a single governing joint distribution
$D_{\cX,\cYc,\cYt}$ for tuples $(x,y_\c,y_\t)$,
though we have no direct access to it,
nor do we observe such tuples.
Rather, for a given example $x \sim D_\cX$
drawn from the marginal distribution,
we observe either the control variable
$y_\c \sim D_{\cYc | \cX=x}$
or the treatment variable $y_\t \sim D_{\cYt | \cX=x}$,
drawn from their respective conditional distributions.

In the setting we consider,
we are given as input three sample sets
of example-label pairs:
\begin{enumerate}
\item 
$S_\c = \left\{ (x^{(i)},y^{(i)}_{\c}) \right\}_{i=1}^M$
sampled i.i.d. from $D_{\cX, \cYc}$
\item
$S'_\c = \left\{ (x^{(i)},y^{(i)}_{\c}) \right\}_{i=1}^{M_\c}$
sampled i.i.d. from $D_{\cXp, \cYc}$
\item
$S'_\t = \left\{ (x^{(i)},y^{(i)}_{\t}) \right\}_{i=1}^{M_\t}$
sampled i.i.d. from $D_{\cXp, \cYt}$
\end{enumerate}

The first set $S_\c$ is a large sample of the general population
with labels for the control variable,
representing past data accumulated by running the default policy.
The sets $S'_\c$ and $S'_\t$ are smaller and represent the results of the
controlled trial for the control and treatment groups.\footnote{
Note that the i.i.d. assumption mimics the procedure of random subject assignment found in RCTs.}
We therefore assume that $M_\c,M_\t \ll M$ 
and that $M_\c \approx M_\t$.
More concretely, we assume that while $M$ is sufficiently
large to learn a reasonably accurate predictor
for the control variable,
$M_\t$ is insufficiently small for
adequate learning of the treatment variable.
Note that we are not guaranteed to have any $x$ for which
we observe both $y_\c$ and $y_\t$.
This is a fundamental problem in counterfactual settings,
and makes estimating the individual treatment effect
$y_\d = y_\t - y_\c$
especially challenging \cite{weiss2015machine}.

\begin{figure}[t]
  \begin{center}
  \includegraphics[width=0.45\textwidth]{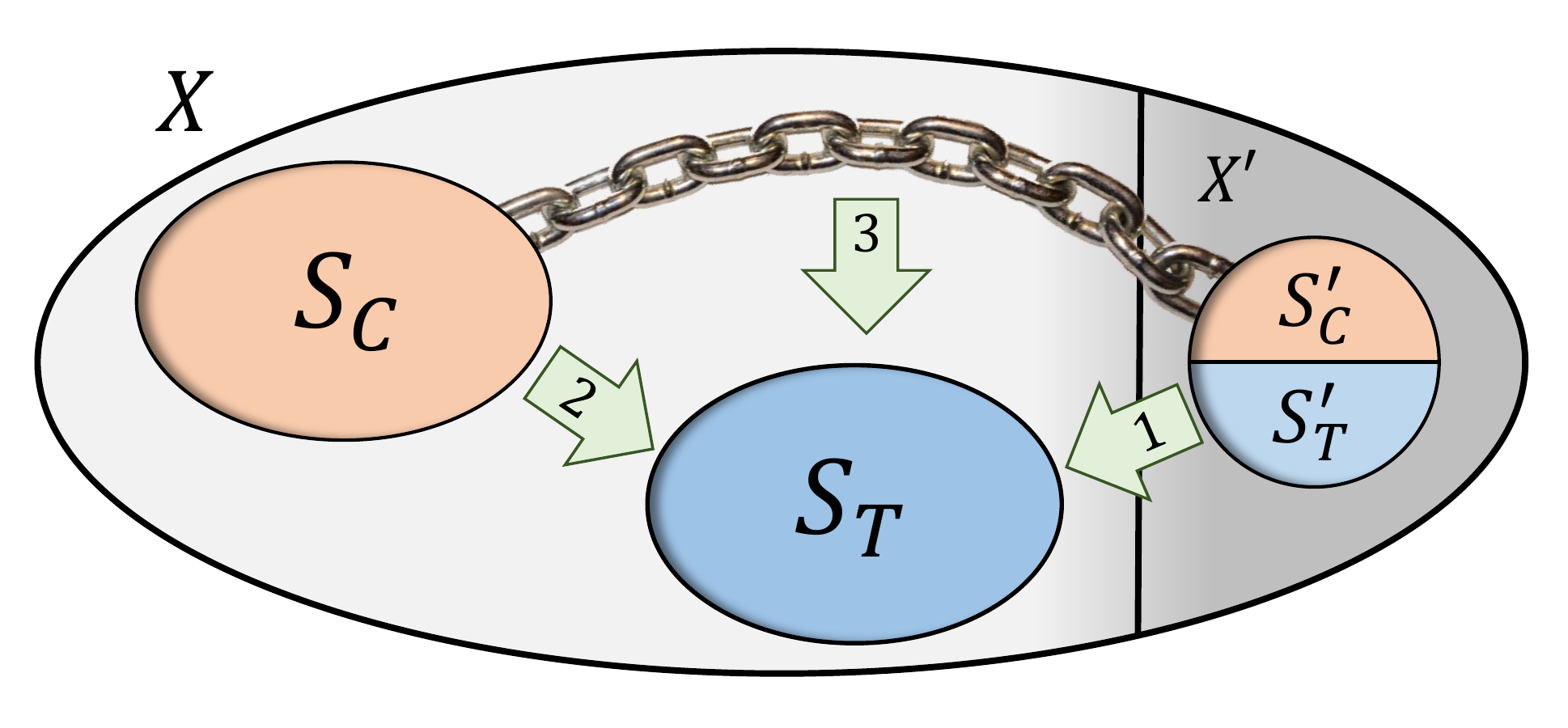}
\end{center}
\caption{
A graphical depiction of the problem setting. Our goal is to predict treatment outcomes for a 
general-population test sample $S_\t$ using:
(1) a small randomized controlled trial from
a (possibly biased) sub-population $X'$, where either of the two possible treatments are randomly given, creating two subsets, $(S'_\c,S'_\t)$, and
(2) a large historical dataset of control outcomes $S_\c$.
Our method links the two datasets using a minimal set of modeling assumptions (3).
}
\label{fig:diagram}
\end{figure}

\section{Problem Statement} \label{sec:problem}
Recall that our goal is to construct a framework
for learning predictors by leveraging both the
small randomized trial data $(S'_\c,S'_\t)$
and the large control-labeled dataset $S_\c$.
The main task we consider is predicting the treatment
variable $y_\t$ for new, unobserved examples
from a test set $S_\t \sim D_{\cX,\cYt}$.
In other words, we'd like our predictor to generalize
well to $\cYt$ on the general population.
The challenge here is that our data contains only
a small number of treatment labels.
The solution we present in Sec. \ref{sec:method} utilizes all the available data
by modeling the relation between the control and treatment variable.

A related task that is of high interest is to predict the
individualized treatment effect $y_\d=y_\t-y_\c$
\cite{rubin1974estimating,van2007causal}.
Accurate predictions of $y_\d$ can in principle aid
decision makers in deciding what treatment to apply.
Such predictions can also be used
to estimate the mean treatment effect $\expect{}{y_\d}$,
and by so offer an alternative to conventional
estimators used in randomized trials.
As we show in Sec. \ref{sec:method},
the individualized treatment effect $y_\d$
plays a central role in our learning objective
for all tasks we consider.


\section{Method} \label{sec:method}

At the core of our method lie only two simple modeling assumptions:
that predictions for both conditions are of the form
$\hat{y}=\inner{w,x}$,\footnote{
In general we assume that predictors are linear in some feature representation $\varphi(x)$,
as we describe in Sec. \ref{sec:ext_linear}.}
and that the models $w_\c,w_\t$ for the control and treatment
conditions, respectively, should be similar under some notion.

For ease of exposition, 
consider first a regression task where $x \in \R^d$, $y_\c,y_\t \in \R$,
and our goal is to minimize the squared loss of a
linear predictor for the treatment variable, namely
$\hat{y}_\t = \inner{w_\t,x}$.
In Sec. \ref{sec:extensions} we show that
our method applies to both regression and classification,
and to a wide of loss functions, and to some
non-linear predictors as well.

Since our task is to predict the treatment outcome $y_\t$
of a given sample $x$,
a reasonable place to start would be in considering a
learning objective over the sample set $S'_\t$,
as it is the only one for which we have treatment labels.
Applying the squared loss
and adding $\ell_2$ regularization gives us:
\begin{align}
\min_{w_\t \in \R^d} \frac{1}{M_\t}
\sum_{i \in S'_\t} \left( \inner{w_\t,x^{(i)}}-y^{(i)}_\t \right)^2
+ \lambda \|w_\t\|_2^2
\label{eq:treatment_loss}
\end{align}

As in any discriminative objective, the number of
samples greatly effects the quality of generalization
of the learned predictor.
Unfortunately, for the above objective and
under our assumptions,
$S'_\t$ will not prove to be sufficiently large for
training a high-accuracy predictor for the treatment variable.
Put simply, our data does not include enough labeled
instances from $\cYt$.

Our approach remedies this deficiency by
artificially augmenting $S'_\t$ with samples that serve as a proxy
for treatment labels. As a first step, we will add to
the objective in Eq. \eqref{eq:treatment_loss}
the samples from $S_\c$, our largest available dataset:
\begin{align}
\min_{w_\t \in \R^d} 
\frac{\gamma}{M_\t} & \sum_{i \in S'_\t} \left( \inner{w_\t,x^{(i)}}-y^{(i)}_\t \right)^2 + \nonumber \\
\frac{(1-\gamma)}{M} & \sum_{i \in S_\c} \left( \inner{w_\t,x^{(i)}}-y^{(i)}_\c \right)^2
+\lambda \|w_\t\|_2^2
\label{eq:t_and_c_loss}
\end{align}
where $\gamma \in  [0,1]$ controls the relative weight of
each dataset in the training objective.
For ease of notation, we overload
$S_\c$ to include all of the available control condition examples,
namely $S'_\c \subset S_\c$.

At a first glance using control outcomes $y_\c$
when trying to predict the treatment outcome $y_\t$ may seem peculiar.
Nonetheless, work in multi-task learning has shown that
training a single predictor over several labels
is beneficial in practice when the
conditional distribution of different labels is similar.
\cite{bickel2008multi}.
However, even if the control and treatment distributions
do share similarities, our focus here is on their differences.
We therefore do not suffice with Eq. \eqref{eq:t_and_c_loss},
in place of the control labels $y^{(i)}_\c$
use proxy treatment labels $\proxy{y}^{(i)}_\t$
which we define next.

Denote by $\D = y_\c-y_\t$ the negative of the
individual treatment effect $y_\d$,
namely the difference between the control and treatment variables.
We have already set $\hat{y}_\t$ to be a linear function of $x$
with weights $w_\t$; extending this to $\hat{y}_\c$ with weights $w_\c$
gives us:
\begin{equation}
\hat{\D} = \inner{w_\c,x} - \inner{w_\t, x} = \inner{w_\d, x}
\label{eq:delta}
\end{equation}
where we use $w_\d=w_\c-w_\t$.
This implies that $\Delta$ is also modeled by
a linear function, and readily gives us our proxy:
\begin{equation}
\proxy{y}_\t = y_\c - \inner{w_\d,x}
\label{eq:yt_proxy}
\end{equation}
Note that this derivation is possible due to our view
of the tuple $(x,y_\t,y_\c)$ as jointly distributed.
This is in contrast to the more conventional approach
where the distribution is modeled using 
tuples of the form $(x,\nu,y_\nu)$, where
$\nu \in \{\c,\t\}$ is the experimental condition
and $y_\nu$ is the outcome under that condition
\cite{johansson2016learning}.
Our formulation induces a joint distribution over pairs $(x,\D) \sim D_{\cX,\D}$, which we can model.

Plugging back into Eq. \eqref{eq:t_and_c_loss}
and further regularizing gives:
\begin{align}
\min_{w_\t,w_\d} 
& \frac{\gamma}{M_\t} \sum_{i \in S'_\t} \left( \inner{w_\t,x^{(i)}}-y^{(i)}_\t \right)^2  + \nonumber \\
& \, \frac{(1-\gamma)}{M}  \sum_{i \in S_\c} \left( \inner{w_\t,x^{(i)}}-\proxy{y}^{(i)}_\t \right)^2  + \nonumber \\
& \, \lambda \|w_\t\|_2^2
\, + \, \eta R(w_\d)
\label{eq:joint_loss_ridge}
\end{align}
where $R$ is a regularization function,
and $\gamma,\eta \in \R$
are additional meta-parameters which we will
shortly describe.
Note that $\proxy{y}_\t$ is in fact a function of $w_\d$;
the explicit form of summands in the second loss term is
$\left( \inner{w_\t-w_\d,x}-y_\c \right)^2$.

To gain insight into the above construction,
we next analyze the learning objective under an
alternative formulation.
Notice that by Eqs. \eqref{eq:delta} and \eqref{eq:yt_proxy},
the second loss term 
and the additional regularization term
in Eq. \eqref{eq:joint_loss_ridge}
can equivalently be written as:\footnote{
This is similar to the regularization term of
the Fused Lasso approach \cite{tibshirani2005sparsity}
used for time-series prediction.}
\begin{equation*}
\sum_{i \in S_\c} \left( \inner{w_\c,x^{(i)}}-y^{(i)}_\c \right)^2, \qquad
\eta R(w_\t-w_\c)
\end{equation*}
Under this representation,
the choice of $R$ and $\eta$ respectively determine
the nature and magnitude of similarity between $w_\t$ and $w_\c$.
For instance, setting $R=\| \cdot \|_2^2$
will encourage $w_\t$ and $w_\c$ to be close under a Euclidian metric, while setting $R=\| \cdot \|_1$ will induce
sparsity on $w_\d$, meaning that $w_\t$ and $w_\c$ will
be different only on a small subset of entries.

This gives an intuitively interpretation of our assumption
on the similarity of $w_\c$ and $w_\t$ via $w_\d$;
we assume that $w_\c$ models the baseline effect,
while $w_\d$ models the deviation of the treatment effect
as expressed by $w_\t$.
This aligns well with our setup.
Since $S_\c$ is large, it should allow for a good fit to the baseline effect of the control condition.
Given this, the fewer samples in $S'_\t$ should now suffice
to fit the deviated treatment effect.
This is especially true for high-dimensional,
where learning requires a large number of samples.
We will return to this in Sec. \ref{sec:ext_nonlinear};

The value of $\eta$ sets the de-facto linkage strength of
the two loss terms in Eq. \eqref{eq:joint_loss_ridge}.
Setting $\eta=0$ will allow $w_\c$ to be arbitrarily
far away from $w_\t$, which will lead to a disjoint objective -
minimizing $w_\t$ over $S'_\t$ and $w_\c$ over $S_\c$ independently.
On the other hand,
setting $\eta=\infty$ will constrain $w_\t=w_\c$ 
and hence revert the objective back to
Eq. \eqref{eq:t_and_c_loss}.

While $\eta$ controls the relation between $w_\t$ and $w_\c$,
$\gamma$ signifies the importance of each sample set
for training $w_\t$ to generalize well to the treatment variable.
While $S'_\t$ contains actual treatment labels but is small, 
$S_\c$ is sufficiently large but contains only control labels
(used as proxies for the treatment variables).
The purpose of $\gamma$ is therefore to allow us to
balance these complementary properties.
Setting $\gamma=1$ will revert the objective back to
Eq. \ref{eq:treatment_loss},
while setting $\gamma=0$ will result in a training objective
based only on $S_\c$.
In effect, the above notions model our belief in how
(and how well) $\proxy{y}_\t$ serves as a proxy for $y_\t$.


\section{Extensions} \label{sec:extensions}
In the above section, we presented our method for
a regression task under a squared loss function
and an $\ell_2$ regularization term.
Note however that our only modeling assumption was 
that both $y_\t$ and $y_\c$ 
(and accordingly $y_\d$)
admitted to linear
predictors under some joint feature representation.
This simple assumption allows us to apply our method
to more general settings, provide a closed-form
solution for some cases.

\subsection{Linear predictors} \label{sec:ext_linear}
An immediate conclusion from the above is that our method
applies to any general loss function $L(\inner{w,x},y)$ defined over a linear predictor, and to any regularization term $Q(w)$ of the predictor's parameters.
The general form of the training objective 
in Eq. \eqref{eq:joint_loss_ridge}
for linear predictors is given by:
\begin{align}
\min_{w_\t,w_\d} 
& \frac{\gamma}{M_\t} \sum_{i \in S'_\t}
L\left( \inner{w_\t,x^{(i)}}, y^{(i)}_\t \right)  + \nonumber \\
& \, \frac{(1-\gamma)}{M}  \sum_{i \in S_\c}
L\left( \inner{w_\t-w_\d,x^{(i)}},y^{(i)}_\c \right)  + \nonumber \\
& \, \lambda Q(w_\t)
\, + \, \eta R(w_\d)
\label{eq:joint_loss}
\end{align}

As we do not make assumptions regarding the nature
of the labels,
Eq. \eqref{eq:joint_loss} is not restricted to regression,
and hence directly applies to binary classification.
For an appropriate definition of $y_\d=y_\t-y_\c$,
Eq. \eqref{eq:joint_loss}  can also be applied in principle
to multi-class and multi-label classification and to structured prediction.
However, note that in such classification settings,
the interpretation of $y_\d$
as the individual treatment effect no longer holds.
For instance, for a margin-based optimization approach
for binary classification,
$y_\d$ signifies the difference in distances to the margin,
rather than the difference in the actual outcome.
For other tasks the role of the regularization term $R$ may also change.

\subsection{Closed form solution}
\label{sec:closed_form}
When applying our method to ridge regression (as in the example in Sec. \ref{sec:method}), 
setting $R(\cdot)=\|\cdot\|_2^2$
allows for a closed form solution
of the objective in Eq. \eqref{eq:joint_loss_ridge}.
This is accomplished by transforming the objective into
a canonical ridge regression form:
\begin{equation}
\min_w \|w^\top X-Y\|_2^2 + \alpha \|w\|_2^2
\label{eq:ridge_obj}
\end{equation}
for which the solution is:
\begin{equation}
\hat{w} = (X^\top X + \alpha I)^{-1} X^\top Y
\label{eq:ridge_sol}
\end{equation}
We now show how to construct the data matrix $X$,
label vector $Y$, and regularization constant $\alpha$,
so that the minimizer of Eq. \eqref{eq:joint_loss_ridge} can be
extracted from $\hat{w}$.

Since the objective in Eq. \eqref{eq:joint_loss_ridge} includes the minimization over both $w_\t$ and $w_\d$,
we first set
$w$ to be their concatenation, namely
$w = (w_\t,w_\d) \in \R^{2d}$.
Under this expanded representation, we next set:
\begin{equation}
\begin{array}{rll}
i \in S'_\t: 
& X_{i \cdot} = c_1 \cdotp (x^{(i)},\,\textbf{0}) 
& Y_i = c_1 y_\t^{(i)} \\
i \in S_\c: 
& X_{i \cdot} = 
c_2 \cdotp (x^{(i)},-c_3 x^{(i)}), 
& Y_i = c_2 y_\c^{(i)}
\end{array}
\label{eq:representation}
\end{equation}
where $\textbf{0}$ is a vector of zeros of size $d$,
and the constants are:
\[
c_1=\sqrt{\gamma/M_\t}, \quad
c_2=\sqrt{(1-\gamma)/M}, \quad
c_3=\sqrt{\lambda / \eta}
\]
Finally, letting $\alpha=\lambda$ and plugging into
Eq. \eqref{eq:ridge_sol} gives the solution for $w_\t$
and $w_\d$ of our original objective in Eq. \eqref{eq:joint_loss_ridge}.

\begin{table}[t]
  \centering
	\small
        \begin{tabular}{|l|l|cccc|c|}
\cline{3-7}    \multicolumn{1}{c}{Task} & \multicolumn{1}{c|}{Measure} & $S'_\t$ & $S_\c$ & $S_{\t' \cup \c}$ & $\Delta$ & $S_\t$ \bigstrut\\
    \hline
    \multirow{2}[2]{*}{\makecell[l]{Stay \\  length}} & mean $r^2$ & 0.153 & 0.068 & 0.167 & \textbf{0.219} & 0.323 \bigstrut[t]\\
          & \% bench. & 47\%  & 21\%  & 52\%  & \textbf{68\%} & 100\% \bigstrut[b]\\
    \hline
    \multirow{2}[2]{*}{\makecell[l]{Above \\ median}} & accuracy & 0.711 & 0.709 & 0.711 & \textbf{0.725} & 0.749 \bigstrut[t]\\
          & \% bench. & 95\%  & 95\%  & 95\%  & \textbf{97\%} & 100\% \bigstrut[b]\\
    \hline
    \end{tabular}%
		\caption{Results of the prediction and classification tasks on the diabetes treatment dataset. The proposed method ($\Delta$) reaches the highest accuracy, compared to methods which use subsets of the available data.}
		
  \label{tbl:diabetes}%
\end{table}%

\subsection{Non-linear predictors} \label{sec:ext_nonlinear}
While linear predictors are easy to work with
and often work well in practice,
they lack the expressive power that non-linear predictors offer.
As our method is not constrained to a specific representation,
a straightforward way for incorporating non-linearity is via kernels, as we describe next.

In Sec. \ref{sec:closed_form}, the construction in
Eq. \eqref{eq:representation} shows how regularizing
of $w_\d$ can be achieved by a simple expansion of the feature representation.
A similar procedure can be applied to a more general case,
specifically when $R,Q$ decompose and $R=Q$.
For $\gamma=1/2$,\footnote{
General values of $\gamma$ can be incorporated into
losses which support differential sample weights.}
setting the expanded features $\phi(x) = (x,\textbf{0})$ for $x \in S'_\t$
and $\phi(x) = (x,-cx)$ for $x \in S_\c$ with $c=\sqrt{\lambda/\eta}$ allows for $R$ and $Q$ to share a single
constant $\lambda$, and due to decomposability define
a single regularization function over the new expanded model
$\tilde{w} \in \R^{2d}$.

Since above holds for any feature representation $\varphi(x)$,
kernel-supporting methods can be readily applied.
For a linear kernel $K(x,x')=\inner{x,x'}$,
the expanded kernel $\bar{K}$ is:
\begin{align}
\bar{K}(x,x') &= \inner{\phi(x),\phi(x)'} =
g(x,x') \cdotp K(x,x'), \nonumber \\
g(x,x') &=
\begin{cases}
c^2  & x,x' \in S_\c \\ 
1  & \mbox{o.w.} 
\end{cases}
\end{align}
As kernels are closed under addition,
for a general feature representation $\varphi$ we have:
\begin{equation}
\bar{K}(x,x') = g(x,x') \cdotp \inner{\varphi(x),\varphi(x')}
\end{equation}
Hence, our method can be applied to wide class of regularized kernel methods,
such as kernel ridge regression, SVMs, SVRs,
and others.

As the dimension of $\varphi$ typically used in kernels
is very large or even infinite,
they require a considerable number of samples to learn properly.
This is also true for many other non-linear predictors.
This makes using kernels only on the small $S'_\t$ unrealistic,
while applying them to $S_\c$ alone is suboptimal.
As mentioned in Sec. \ref{sec:method},
our method should allow for using the large number of samples
in $S_\c$ to learn the baseline effect of the control condition
using kernels, while still taking advantage of
the treatment-labeled samples in $S'_\t$.

Finally, we note that since many non-linear
deep architectures include
a linear output layer,
our method can potentially be applied to such.
In a similar fashion to the construction in Sec. \ref{sec:closed_form},
such an architecture should include two linear output layers - one for $w_\t$ and one for $w_\d$ - 
and corresponding regularization terms.
We leave the exploration of such an approach for future work.

\vspace{20pt}


\section{Experiments} \label{sec:experiments}

\begin{table}[t]
  \centering
	\small
    \begin{tabular}{|l|l|cccc|c|}
\cline{3-7}    \multicolumn{1}{c}{Task} & \multicolumn{1}{c|}{Measure} & $S'_\t$ & $S_\c$ & $S_{\t' \cup \c}$ & $\Delta$ & $S_\t$ \bigstrut\\
    \hline
    \multirow{2}[2]{*}{Value} & Mean $r^2$ & 0.564 & 0.651 & 0.660 & \textbf{0.688} & 0.716 \bigstrut[t]\\
          & \% bench. & 79\%  & 91\%  & 92\%  & \textbf{96\%} & 100\% \bigstrut[b]\\
    \hline
    \multirow{2}[2]{*}{\makecell[l]{Top \\ decile}} & Accuracy & 0.849 & 0.831 & 0.853 & \textbf{0.861} & 0.875 \bigstrut[t]\\
          & \% bench. & 97\%  & 95\%  & 97\%  & \textbf{98\%} & 100\% \bigstrut[b]\\
    \hline
    \end{tabular}
		\caption{Results of the prediction and classification tasks on the housing dataset. The proposed method ($\Delta$) reaches the highest accuracy and 96\% or more of the benchmark ($S_T$), which uses data from the entire treatment dataset.}
  \label{tbl:houses}%
\end{table}%

\begin{table*}[t]
  \centering
\begin{tabular}{llcccccccccc}
\cline{3-12}    \multicolumn{1}{c}{Task} & \multicolumn{1}{c|}{Measure} & $S'_\t$ & $S_\c$ & $S_{\t' \cup \c}$ & \multicolumn{1}{c|}{$\Delta$} & \multicolumn{1}{c||}{$S_\t$} & $S'_\t$ & $S_\c$ & $S_{\t' \cup \c}$ & \multicolumn{1}{c|}{$\Delta$} & \multicolumn{1}{c|}{$S_\t$} \bigstrut\\
    \hline
    \multicolumn{1}{|l|}{\multirow{2}[2]{*}{\makecell[c]{Individual treatment \\ outcome $y_\t$}}} & \multicolumn{1}{l||}{mean $r^2$} & -0.04 & 0.19  & 0.21  & \multicolumn{1}{c|}{\textbf{0.26}} & \multicolumn{1}{c||}{0.30} & 0.26  & 0.22  & 0.24  & \multicolumn{1}{c|}{\textbf{0.33}} & \multicolumn{1}{c|}{0.36} \bigstrut[t]\\
    \multicolumn{1}{|l|}{} & \multicolumn{1}{l||}{\% bench.} & -     & 65\%  & 72\%  & \multicolumn{1}{c|}{\textbf{89\%}} & \multicolumn{1}{c||}{100\%} & 72\%  & 63\%  & 68\%  & \multicolumn{1}{c|}{\textbf{94\%}} & \multicolumn{1}{c|}{100\%} \bigstrut[b]\\
    \hline
    \multicolumn{1}{|l|}{\multirow{2}[2]{*}{\makecell[c]{Individual treatment \\ effect $y_\d$}}} & \multicolumn{1}{l||}{mean $r^2$} & -0.08 & 0.18  & 0.20  & \multicolumn{1}{c|}{\textbf{0.24}} & \multicolumn{1}{c||}{0.26} & 0.22  & 0.22  & 0.23  & \multicolumn{1}{c|}{\textbf{0.31}} & \multicolumn{1}{c|}{0.32} \bigstrut[t]\\
    \multicolumn{1}{|l|}{} & \multicolumn{1}{l||}{\% bench.} & -     & 70\%  & 76\%  & \multicolumn{1}{c|}{\textbf{90\%}} & \multicolumn{1}{c||}{100\%} & 69\%  & 67\%  & 72\%  & \multicolumn{1}{c|}{\textbf{95\%}} & \multicolumn{1}{c|}{100\%} \bigstrut[b]\\
    \hline
    \hline
    \multicolumn{1}{|c|}{Average effect $\expect{}{y_\d}$} & \multicolumn{1}{l||}{abs. diff.} & 0.23  & 0.27  & 0.24  & \multicolumn{1}{c|}{\textbf{0.14}} & \multicolumn{1}{c||}{0.06} & \textbf{0.05} & 0.24  & 0.23  & \multicolumn{1}{c|}{0.07} & \multicolumn{1}{c|}{0.05} \bigstrut\\
    \hline
          &       & \multicolumn{5}{c}{25:75 split}       & \multicolumn{5}{c}{75:25 split} \bigstrut[t]\\
    \end{tabular}%
\caption{
		Search engine ranking results for predicting individualized treatment outcomes and effects (higher is better)
		and the average treatment effect for an A/B test (lower is better) using different sample training sets.
		The proposed method ($\Delta$) links both samples by enforcing similarity.
		Results are averaged over all A/B tests.
		$S_\t$ is used as a high-end benchmark.
		}
  \label{tbl:flights}%
\end{table*}%


In this section we evaluate the performance of our method
on three counterfactual prediction tasks: A simulated medical clinical trial, a web search engine experiment, and a social choice question.
Since our learning goals include predictions regarding the treatment variable $y_\t$,
our data must contain a large pool of ground-truth labels for this class.
This is a necessary condition for ensuring a valid estimation procedure.
Unfortunately, for the same reasons that motivate our work,
most datasets do not include many labeled treatment instances,
as they are typically hard, expensive, and time consuming to acquire.

To this end, we focus on three datasets.
The first dataset contains information on the clinical status of approximately 100,000 diabetes patients. Our task is to predict the length of hospitalization for each patient, given their treatments so far. 

A second dataset comprises of a large collection of around 20,000 houses along with their attributes.
Our task is to estimate the market price of a house given its attributes.
While the dataset itself was not collected by a randomized trial procedure,
we partition the recordings into control and treatment conditions
in a way which emulates a realistic controlled trial scenario.
This allows us to validate our predictions on the treatment variable.

The third dataset is from the domain of search engine operation.
Search engines regularly modify and improve their ranking algorithm and other parameters such as the user interface,
in many cases based on the results of a large number of A/B tests,
where the current ranking algorithm is compared to a new alternative. In this setting, early and accurate predictions of query-centric measures like the \emph{click-through rate} (CTR) and its derivatives are of great importance. Thus, we focus on this prediction, which can also assist in early termination of treatments which are predicted to be as good as (or worse than) the current treatment.

At its core, our method provides a way to model the linkage
between a small randomized trial
and a large historical dataset.
Our goal in this section is therefore to evaluate
the added benefit of using our model when such data is available.
In Sec. \ref{sec:extensions}, we describe why and how our method can be applied to a large set of loss functions and predictors.
To this end, in this section we compare the performance of our method
to methods which simply aggregate both datasets, while keeping the loss and predictor class fixed.
Specifically, we compare our method to training only on $S'_\t$,
only on $S_\c$,
and on both sets $S_{\t' \cup \c} = S'_\t \cup S_\c$.
As for other linear methods described in Sec.
\ref{sec:related},
\cite{swaminathan2015counterfactual,swaminathan2015self}
assume that samples include loss terms and propensity scores,
while the linear method in \cite{johansson2016learning}
does not outperform standard ridge regression.

We evaluated performance on two tasks:
predicting {\bf individual treatment outcomes} ($y_\t)$,
and predicting the {\bf individual treatment effect} ($y_\d$).

As mentioned, all methods were evaluated on a held-out test set with treatment labels,
were results were averaged over 10 random instantiations.
For all tasks we used ridge regression as a learning objective,
and applied $\ell_2$ regularization for our method.
Meta-parameters were chosen on a small held-out validation set.
Our high-end benchmark for performance is based on learning over
a large treatment-labeled training set $S_\t$.
We note again that this type of data is typically unrealistic to attain,
and hence serves as an empirical upper bound on overall predictive accuracy.

\subsection{Hospitalization of diabetes patients}

This dataset contains data from 10 years (1999-2008) of clinical care at 130 US hospitals and integrated delivery networks \cite{strack2014impact}. We attempt to predict the length of hospitalization (in days), or (for the classification task) whether the length of hospitalization would be longer than the median hospitalization length. 

The 'new' treatment which we attempt to estimate is whether prescription of diabetes medications prior to hospitalization would have changed the hospitalization length.  
We focused on patients for which the reason of admission
was unknown, as these represent the difficult cases,
of which there were 4,785 patients in the data.
We simulated a clinical trial by randomly selecting 25\% of the population into $\cX'$, some of whom were prescribed diabetes medications, and some who were not.
We used 25:75 train-test splits.

We evaluate performance on two tasks:
predicting the hospitalization length,
and predicting whether the length was above the median.
The results of these experiments are shown in Table \ref{tbl:diabetes}. As the results show, our methods provides better estimation of the treatment effect, compared to methods which are based on subsets of available data. Moreover, this prediction is close in its quality to that achieved by a learner which uses the actual treatment information.

\subsection{House pricing Dataset}
The House Sales in King County dataset\footnote{\url{https://www.kaggle.com/harlfoxem/housesalesprediction}}
contains records of 21,613 houses sold in King County, USA,
a region which includes Seattle.
Along with the market price of each house,
the data includes 19 numerical and categorical attributes
for each house including
the number and types of rooms, size, number of floors,
and geographic location.
Of special interest is an attribute which determines
whether the house was renovated or not.
By considering this as a treatment indicator variable
and partitioning accordingly,
we can simulate the following counterfactual question:
Does renovating increase a house's value, and by how much?

As houses were not randomly assigned to each condition,
the data does not represent a true randomized controlled trial.
This of course raises questions as to whether predictions
can be used to answer the above question.
Nonetheless, our method still applies here,
as we do not assume a random assignment,
but rather use it as motivation.

We evaluated performance on two tasks:
a regression task in which we predict the value of a house,
and a classification task in which we predict
whether the value is in the top decile.
In both tasks we assign all houses which did not undergo renovation
to the control condition,
and all those which did to the treatment condition.
These amounted to 20,699 and 914 houses, respectively.
We used 75:25 train-test splits,
and set the experimental sub-population $\cX'$ to be all houses
in a  random subset of zip codes, representing about
25\% of all zip codes.
This is similar to a setting where only certain areas
residential areas participate in a survey.

Results for all tasks
are presented in Table \ref{tbl:houses}.
We report the mean $R^2$
for the task of predicting a house's value,
and mean accuracy for predicting the attribution to the top decile.
Results show that the highest accuracy in both prediction tasks is obtained by the proposed method (denoted as $\Delta$ in the table). Moreover, these accuracies are within a few percentage points from the accuracy obtained when using data from benchmark dataset. Thus, the proposed method can replace the use of a large RCT, which would be expensive and difficult to execute, in this kind of setting.

\subsection{Search Engine Dataset} \label{sec:flights}

We collected queries submitted to the Bing search engine on June $1^{\text{st}}$ 2016
which were randomly assigned to internal A/B tests, and whose frequency was at least 1,000.
As examples ($x$) we took all of the queries that appeared in the control condition,
and focused on tests which included all of these queries.
Our dataset contains 277 comparable treatment conditions and one control condition, each with 1,572 distinct query examples,
for a total of 437,016 instances.

Features included both categories of the queries and features of the query words. Categorization was determined using a proprietary classifier \cite{yom2014seeking} developed by the Microsoft Bing team to assign each query into a set of 63 categories, including, for example, commerce, tourism, video games, weather-related, and adult-themed queries. The classifier is used by Bing to determine whether to display special results such as instant answers. Queries can be classified into multiple categories (e.g., purchase of flight tickets would be classified into both tourism and commerce). 
Word-based features included some basic attributes such as
the number of characters,
number of tokens, minimal and maximal token size,
and a numeric token indicator.
In addition, a bag-of-words representation of the tokens
was computed, and applied using the feature hashing trick
\cite{weinberger2009feature}.

To generate labels, for each query $x$ and for each experimental condition,
we computed CTR estimates by pooling all relevant query instances.
Since the distribution of CTR is highly skewed, and since some queries have CTR$=0$,
we set label values to $y=\log(\text{CTR}+\epsilon)$ for $\epsilon=10^{-10}$.
The above process ensured that for a given $x$ our data included both $y_\c$ and $y_\t$, from which $y_\d$ was computed.

A distinct characteristic of this dataset is that
for most search phrases $x$, the data includes both
$y_\c$ (the CTR under the default ranker)
and $y_\t$ (the CTR under the new ranker).
This is because, for common search phrases,
responses will be recorded under both control and treatment conditions.
The above allows us to directly evaluate the individual treatment effect $y_\d$.
Moreover, since the data includes a large collection of alternative rankers 
(but only one default ranker),
we can compare the effect of many treatments to the same control condition.

Running search engine A/B tests is an expensive procedure.
Tests are therefore often limited in time and resources,
and are allocated only a small fraction of the overall traffic.
This causes CTR estimates to be unrepresentative,
as high-frequency queries will be assigned to an A/B test
more often than low-frequency queries,
which may not appear in trials at all.
Therefore, our proposed algorithm can potentially shorten A/B test by reaching a conclusion as to the benefit of a new ranking algorithm using only the popular queries (which are easy to collect), but inferring the benefit for rare queries as well, solely based on a complete historical record of the control condition and a small random trial over a non-representative  population (e.g., popular queries).

Since our estimation procedure requires a full set
of ground-truth treatment labels,
we can use only query instances which participated in trials.
This requires us to mimic the above setting using A/B test data alone,
by constraining $S'_\t$
to contain only queries whose frequency
is in the top $\tau^{\text{th}}$-quantile.
For instance, by setting $\tau=0.75$, we guarantee that $S'_\t$
contains only queries with frequency in the top quartile.
To keep $S'_\t$ small, we further discard a random $25\%$ of the qualified examples.

For comparison of the proposed method we use the results of learning on $S_\t$ with $\tau=0.75$. This means learning on most of the available data, necessitating a long A/B test to collect the rarer queries. We refer to this learning task as the benchmark.

Results for all prediction tasks appear in Table \ref{tbl:flights}.
For individualized treatment outcomes and effects we report the mean $R^2$ and its fraction of the benchmark. As can be seen, our method significantly outperforms learning using the available subsets of the data by a significant margin. Indeed, the accuracy of the proposed algorithm is not far from that of the benchmark, reaching approximately 90\% or more of the potential accuracy for both tasks.

\begin{figure}[t]
  \begin{center}
  \includegraphics[width=0.9\columnwidth]{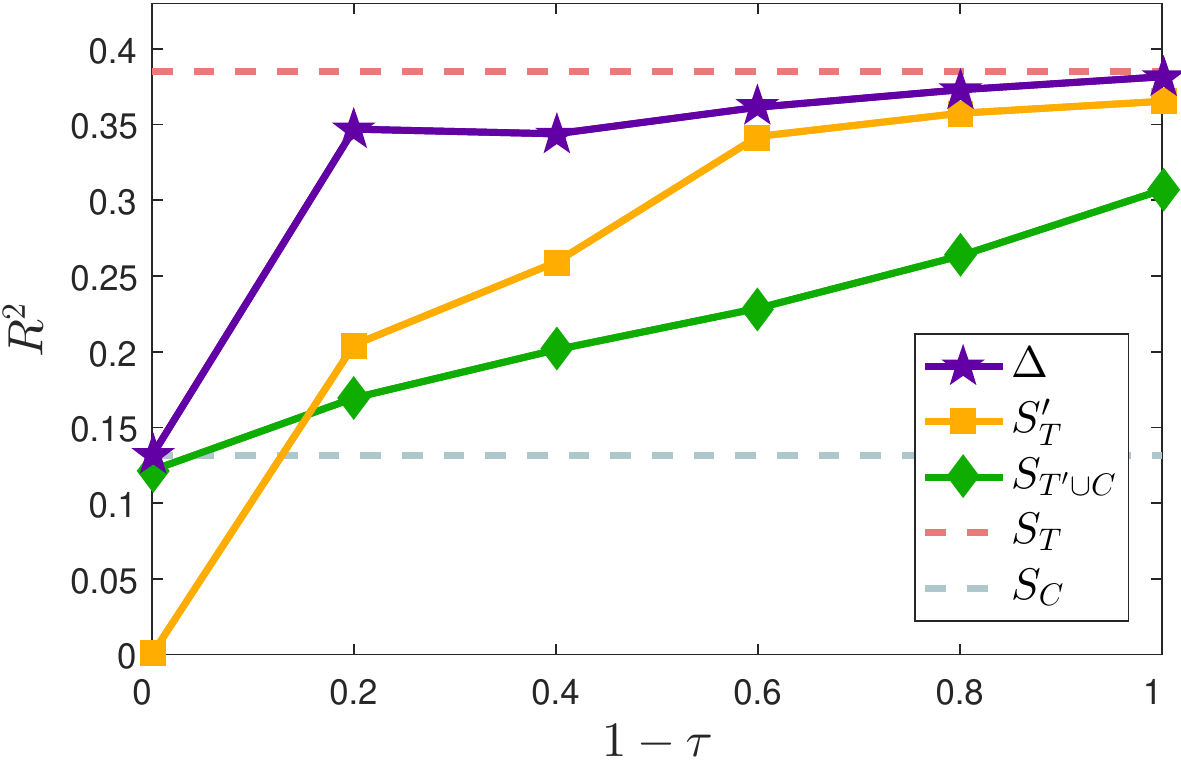}
\end{center}
\caption{Accuracy of the regression algorithm as a function of the fraction of queries used for training, by datasets used for training. As the fraction of queries increases ($\tau$ decreases), $S'_\t$ includes more queries.
This mimics a setting where the length of an A/B test
determines the frequency threshold of observed items. The proposed method ($\Delta$) quickly reaches performance close to that of the benchmark, which uses the treatment information for all queries.
}
\label{fig:flights_time}
\end{figure}

To explore the effect of trial duration
(represented here by thresholding frequency),
we repeated the above procedure for various values of $\tau$.
To accentuate results, we focus on the the top 10\% of trials for
which the difference between conditions was a-priori most significant,
and employed a 50:50 train-test ratio.
Results are presented in Figure \ref{fig:flights_time}. Our method enjoys a fast growth rate in accuracy,
and should potentially allow for shorter trial lengths, or for early stopping of trials, when the new treatment is deemed to be inferior to existing treatments.

\section{Discussion} \label{sec:discussion}

Randomized controlled trials (RCTs) are the gold standard for testing new treatments and interventions. RCTs are widely used by Internet websites, by medical authorities, and, increasingly, by governments. However, RCTs are difficult and expensive to run. Nowadays, historical data is available in many settings where RCTs are considered. However, as these data were collected using an existing policy, utilizing these data has proven difficult. 

In this paper we proposed a new algorithm for using historical data in conjunction with the results of small RCTs, to counterfactually infer the outcomes of large RCTs. Our method can additionally be used as an early stopping criterion for RCTs, when the method predicts that the benefit of a new treatment will not be larger than those of the existing treatment. Thus, our method can provide benefit to existing RCTs.

The proposed method is based on two assumptions: The first is that the outcome of each treatment can be predicted using a linear predictor. The second is that the difference between the  predictor of the current treatment and the proposed treatment is not large.  In Sec. \ref{sec:ext_nonlinear} we showed extensions to the method which overcome the first assumption. Moreover, we hypothesize that predictions can be improved by using robust regression, or through inclusion of a confidence measure for each point in our data. Such extensions are left for future work.

\clearpage

\bibliographystyle{abbrv}
\bibliography{refs}

\end{document}